\documentclass[letterpaper]{article} 
\usepackage{aaai24}  
\usepackage{times}  
\usepackage{helvet}  
\usepackage{courier}  
\usepackage[hyphens]{url}  
\usepackage{graphicx} 
\usepackage{natbib}  
\usepackage{caption} 
\usepackage{algorithm}
\usepackage{algorithmic}

\usepackage{newfloat}
\usepackage{listings}
\DeclareCaptionStyle{ruled}{labelfont=normalfont,labelsep=colon,strut=off} 
\floatstyle{ruled}
\newfloat{listing}{tb}{lst}{}
\floatname{listing}{Listing}
\usepackage{amsmath}
\usepackage{amssymb}

\title{PSGText: Stroke-Guided Scene Text Editing with PSP Module}

\author {
    Felix Liawi,
    Yun-Da Tsai,
    Guan-Lun Lu,
    Shou-De Lin,
}
\affiliations {
    National Taiwan University\\
    felixliawi@gmail.com, f08946007@csie.ntu.edu.tw, allen522019@gmail.com, sdlin@csie.ntu.edu.tw
}

\usepackage{bibentry}
\newcommand{\model}{PSGText}

\begin{document}

\maketitle

\begin{abstract}
Scene Text Editing (STE) aims to substitute text in an image with new desired text while preserving the background and styles of the original text. However, present techniques present a notable challenge in the generation of edited text images that exhibit a high degree of clarity and legibility. This challenge primarily stems from the inherent diversity found within various text types and the intricate textures of complex backgrounds. To address this challenge, this paper introduces a three-stage framework for transferring texts across text images.
Initially, we introduce a text-swapping network that seamlessly substitutes the original text with the desired replacement. Subsequently, we incorporate a background inpainting network into our framework. This specialized network is designed to skillfully reconstruct background images, effectively addressing the voids left after the removal of the original text. This process meticulously preserves visual harmony and coherence in the background. Ultimately, the synthesis of outcomes from the text-swapping network and the background inpainting network is achieved through a fusion network, culminating in the creation of the meticulously edited final image. A demo video is included in the supplementary material.
\end{abstract}

\section{Introduction}
The endeavor to modify text within an image while upholding its inherent style and background constitutes a complex undertaking within the realm of computer vision. Manual interventions by human operators, though effective, tend to consume a considerable amount of time. This intricate task, commonly referred to as Scene Text Editing (STE), has emerged as a subject of noteworthy interest, propelled by its wide-ranging utilities across diverse domains, including the production of advertisements, magazines, and interactive gaming scenarios.

The objective of this work is to develop an algorithm capable of automatically changing the text in an image, thereby streamlining the editing process. The automation of STE can significantly enhance efficiency and productivity in industries that heavily rely on visual content creation, such as advertising, etc. By enabling automated text editing, professionals and designers can focus on higher-level tasks, leading to faster turnaround times and increased creativity in their work.
Moreover, the potential impact of automating STE extends beyond the realm of content creation. It can potentially revolutionize industries that heavily rely on text-based information, such as publishing, advertising, and branding. By efficiently modifying text in images, businesses can quickly adapt their visual materials to cater to changing requirements, target different demographics, or experiment with various design options.
Furthermore, STE maintains a close kinship with other pivotal tasks within the domain of computer vision, notably encompassing scene text detection and scene text recognition. Through the strategic harnessing of STE methodologies, the intricate process of amassing an expansive dataset of scenes for the purposes of training gains a heightened level of feasibility. This unfolding prospect subsequently paves the way for novel opportunities in elevating the efficacy of scene text detection and recognition systems, as they stand to benefit from the incorporation of augmented training datasets characterized by greater scope and diversity.
\begin{figure*}[h]
  \includegraphics[width=\textwidth]{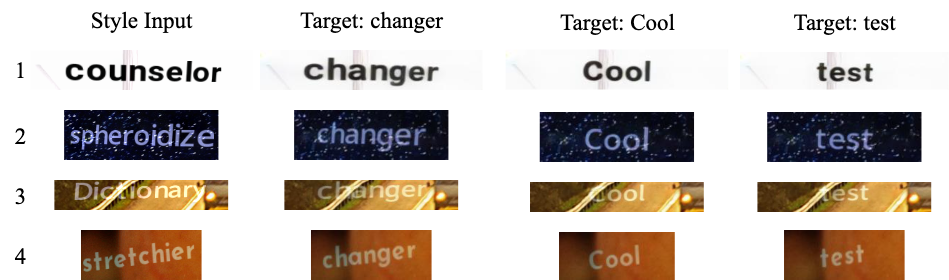}
  \caption{Illustrative examples generated using \model. From left to right: (1) Input image, (2) Model output with content changed to "changer" text, (3) Model output with content changed to "Cool" text, and (4) Model output with content changed to "test" text.}
  \label{fig:results}
\end{figure*}

This paper presents \model, a novel algorithm for Scene Text Editing (STE). It constitutes an end-to-end framework featuring three sub-networks: Background Inpainting, Text Swapping, and Fusion. Leveraging the Pyramid Scene Parsing (PSP) module in conjunction with stroke guidance, our approach enables efficient and accurate transfer of both style and content. By effectively decomposing these elements, our algorithm achieves refined style and content manipulation in scene text editing.
We introduce the Pyramid Scene Parsing (PSP) module as a fundamental component of \model. By integrating PSP with resnet ~\cite{he2016deep}, we successfully decompose the style and content of the text in the images. Our experiments demonstrate the effectiveness of this approach, showcasing SOTA results in style content transfer.
Moreover, \model~benefits from the incorporation of stroke guidance, which proves to be pivotal in accelerating the style transfer process. The stroke guidance mechanism ensures coherence and consistency in the transferred style, resulting in high-quality edited text with minimal distortion.

To further boost the performance, we present a background removal technique inspired by the findings in~\cite{tang2021stroke}. This background removal module serves as one of the three core components of \model. 
The background removal technique effectively eliminates unwanted backgrounds, producing polished and refined text outputs. A few examples can be seen in Figure ~\ref{fig:results}.
Finally, this paper introduces a synthetic dataset generator for scene text editing (STE) research, filling a notable gap in publicly available resources. To the best of our knowledge, this is the first official dataset generator presented in the literature. Our proposed synthetic data generator prioritizes simplicity in configuration, making it accessible and user-friendly. Additionally, it places a strong emphasis on real-world digital image use cases, able to adjust image opacity (non-transparency), which enhances its relevance and practical applicability.

In the following sections, we provide an in-depth analysis of related works, exploring the existing literature on scene text editing and style transfer. By positioning \model~ within this context, we highlight its innovative contributions and underline its potential impact on various applications that involve scene text manipulation and generation.
Our contributions are summarized as follows:
\begin{itemize}
  \item We design an end-to-end framework, namely \model, which contains three sub-networks, text swapping network, background completion network, and the fusion network.
    \item \model~achieves promising performance in both quality and quantity. Our simple and powerful model will provide many insights for future STE works.
    \item We released a novel synthetic dataset generator for scene text editing with strong practical applicability.
\end{itemize}

\section{Related Works}

\subsection{Style Transfer}
Style transfer is a challenging task that involves the transformation of the visual style of a source image to a target image. The majority of existing methods utilize encoder-decoder architectures to generate the target image. \cite{luan2017deep} introduced a deep photo style transfer technique that employs a mapping network to learn a mapping from an input photo to the parameters of a synthesis network, which then generates the stylized output image from a random noise vector. \cite{li2017diversified} proposed a texture synthesis method based on feed-forward neural networks that train a generator network to perform style transfer by adjusting the mean and variance of feature maps and introduces a diversity regularization term to encourage the generation of a wider range of textures. \cite{karras2019style} implemented a mapping network that learns to map vector noise z to weights, which are then fed to a synthetic network to create the target image.

\subsection{Text Image Synthesis}
Text image synthesis has become an increasingly important technique for enhancing the performance of deep neural networks in various text-related tasks, such as data augmentation for text detection and recognition. For example,~\cite{krishnan2016generating} proposed a method for Generating Synthetic Data for Text Recognition, which involves synthesizing text images using a variety of fonts and styles, followed by realistic augmentation techniques such as rotation, scaling, and blurring. They demonstrate that combining the synthetic data with real data leads to improved recognition performance on multiple benchmarks, highlighting the potential of synthetic data as a valuable resource for training text recognition models. SynthTIGER~\cite{yim2021synthtiger}, a synthetic text image generator, which utilizes a Generative Adversarial Network (GAN) to generate synthetic text images and employs a novel embedding-based technique to condition the GAN to generate diverse and realistic images.

\subsection{Scene Text Editing}
Scene text editing presents a formidable challenge as it strives to transfer text style while faithfully preserving the background in a realistic manner. Numerous recent studies in this domain have employed GAN-based models. For instance, SRNet~\cite{wu2019editing} proposes a hierarchical editing approach consisting of three sub-networks: background inpainting, text swapping, and a fusion network. Conversely, STEFANN~\cite{roy2020stefann}, a Font Adaptive Neural Network, albeit with constraints limited to character-level modifications. Notably, ~\cite{qu2023exploring} put forth an innovative network named MOSTEL (MOdifying Scene Text Image at strokE Level) that modifies scene text images at the stroke level.

\begin{figure*}[h]
  \includegraphics[width=\textwidth,height=8cm]{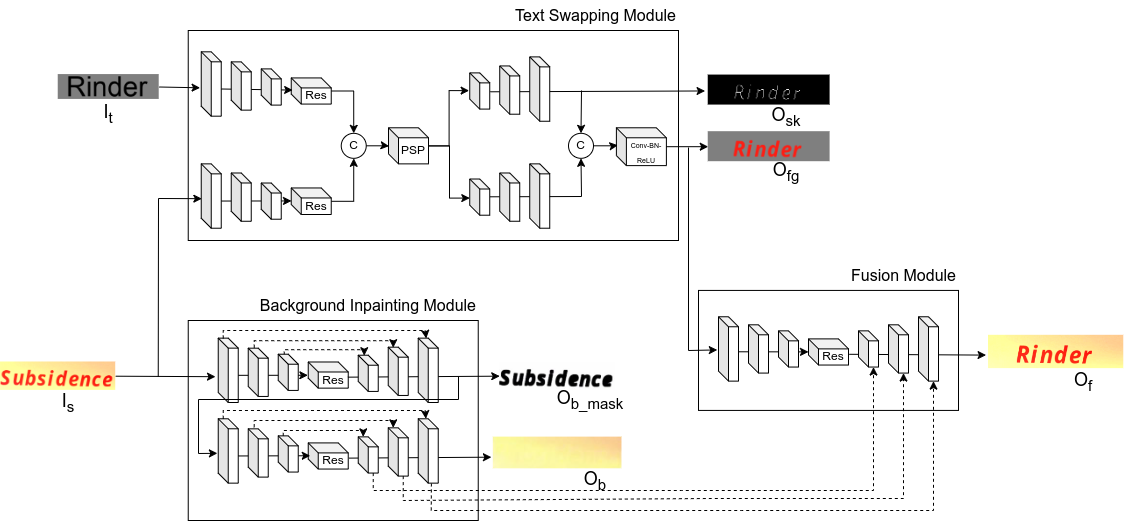}
  \caption{\model~overall structure which comprises the Background Inpainting Module (BIM), Text Swapping Module (TSM), and Fusion Module (FM). The task is to transfer style image Subsidence $I_s$ into content image Rinder $I_t$ and maintain the same style/font resulting $O_f$. }
  \label{fig:whole_pipeline}
\end{figure*}

\section{Methodology}
The proposed methodology aims to generate highly realistic results while ensuring the preservation of the background to the maximum extent possible. This objective is achieved through the utilization of a three-module network. Figure ~\ref{fig:whole_pipeline} illustrates the overall pipeline of the proposed method, which comprises the Background Inpainting Module (BIM), Text Swapping Module (TSM), and Fusion Module (FM).
The BIM module takes the source style image $I_s$ as input and generates the source foreground mask $O_{b\_mask}$ and a clean background $O_b$. Subsequently, the TSM module accepts both the source style image $I_s$ and a target text image or content image $I_t$ in standard format as inputs and produces two transfer results: one in grayscale skeleton form $O_{sk}$ and another in RGB format $O_{fg}$. Finally, the FM module takes the text-free background and the RGB transfer result as inputs and generates the final output $O_f$ of the scene text editing process.
Further details regarding each module will be elaborated upon in the subsequent sections.

\subsection{Background Inpainting Module}

The objective of the Background Inpainting Module (BIM) is to generate a text image devoid of any lingering text artifacts. Inspired by the work of ~\cite{tang2021stroke}, BIM adopts a double encoder-decoder architecture.
The first encoder-decoder architecture, called the Stroke Mask Module (SMM), is utilized to extract the text mask. The second, known as the Text Removal Module (TRM), is responsible for removing the text from the original image using the text mask and the original image as inputs.
The SMM encoder comprises three downsampling blocks and one residual block, while the decoder consists of three upsampling blocks. On the other hand, the TRM encoder consists of three partial convolution downsampling blocks, and one self-attention block, and the decoder consists of three partial convolution upsampling blocks. This architecture allows the TRM module to effectively remove the text while preserving the integrity of the original background image. BIM generates two outputs which are mask output $O_{b\_mask}$ and clean background $O_b$.
To ensure that BIM generates good results, we adopted L2 loss as in Equation~\ref{eq:bi} as the optimization criterion where $T_b$ represents the ground truth for the clean background, and $O_b$ denotes the clean background predicted by the module.
\begin{align}
    L_{b\_i} = \| T_b - O_b \|_2
    \label{eq:bi}
\end{align}
$L_{b\_i}$ penalizes large errors more heavily than small errors. Given the importance of preserving the fine details of the background image during text removal, L2 loss serves as an appropriate choice to encourage the model to produce accurate and visually appealing modifications. Additionally, L2 loss aligns well with the regression objective of BIM, aiming to minimize the pixel-wise differences between the generated and ground-truth background images.

\subsection{Text Swapping Module}

Text instances in our physical world exhibit a diverse range of shapes, encompassing both horizontal and curved forms. The primary aim of the proposed Text Swapping Module (TSM) is to effectively replace the textual content present in a given style image while meticulously preserving the original styles and background. To accomplish this objective, we introduce a Text Swapping Module that facilitates the acquisition of a learned mapping between the content image and the style image. Both the style and content inputs undergo a thorough processing pipeline involving three layers of encoder-decoder networks as well as a ResNet layer. Additionally, the content features are merged with the style features employing a Pyramid Scene Parsing Network~\cite{zhao2017pyramid}, followed by the utilization of a double encoder-decoder network. One of the encoder-decoder networks focuses on generating the skeleton structure denoted as $O_{sk}$, while the other network is responsible for generating the foreground images referred to as $O_{fg}$. $O_{sk}$ is used to achieve faster style transfer by guiding the model to create the desired structure.
To ensure that TSM generates good results, we adopted L2 loss between the predicted text segmentation mask, denoted as $O_{fg}$, and the corresponding original image $T_{fg}$. $L_{ts\_fg}$ shown in Equation~\ref{eq:tsfg} is well-suited for this task as it encourages the model to produce segmentation masks that closely resemble the ground-truth images, emphasizing the preservation of fine details during text generation.

\begin{align}
    L_{ts\_fg} = \| T_{fg} - O_{fg} \|_2
    \label{eq:tsfg}
\end{align}

In addition to L2 loss, we have incorporated the dice loss ~\cite{milletari2016v} to further enhance the quality. The dice loss is applied between the predicted text skeleton, represented as $O_{sk}$, and the corresponding original skeleton image $T_{sk}$. The dice loss shown in Equation~\ref{eq:dice} has been widely used in segmentation tasks and is known for its ability to handle imbalanced datasets, making it suitable for capturing fine text details while maintaining accurate text boundaries in the generated results.

\begin{equation}
    L_{ts\_sk} = 1 - \frac{2 \sum_{i}^{N} (T_{sk})_i (O_{sk})_i}{\sum_{i}^{N} (T_{sk})_i^2 + \sum_{i}^{N} (O_{sk})_i^2}
    \label{eq:dice}
\end{equation}

Lastly, we employ GAN loss, which motivates the TSM to generate text images that are virtually indistinguishable from real images, achieved by setting up a competition between the generator and a discriminator.

The GAN loss is
\begin{equation}
\begin{split}
    L_{ts\_gan} = \mathbb{E}\{\log D_{ts}(T_{fg}, O_{b\_inverted\_mask}) \\ +  \log(1-D_{ts}(O_{fg}, O_{b\_inverted\_mask})\}
\end{split}
\end{equation}

Where $O_{b\_inverted\_mask}$ is the inverted version of $O_{b\_mask}$, achieved by reversing the black and white areas, creating a visual transformation from dark to light and vice versa.

The overall loss function utilized for this module is defined as follows

\begin{align}
    L_{ts} = \lambda_{ts1} L_{ts\_{fg}} + L_{ts\_sk} + L_{ts\_gan} 
\end{align}

Where $\lambda_{ts1}$ is balance factor and is set to 10.

\subsection{PSP Network}

To incorporate the style and content features effectively, we employ a Pyramid Scene Parsing (PSP) module. The PSP module is designed to capture multi-scale contextual information from both the style and content images. It operates by dividing the input feature maps into multiple regions and extracting features at different scales.
In our approach, the style and content features are fed into the PSP module, which consists of parallel convolutional branches with varying receptive field sizes. Each branch captures contextual information at a different scale, enabling a comprehensive understanding of the scene. The resulting feature maps from each branch are then fused together to form a rich representation that encompasses both local and global contexts.
By leveraging the PSP module, our model gains the ability to effectively encode and incorporate the intricate details and contextual cues from both the style and content images. This enables the Text Swapping Module (TSM) to generate accurate and visually appealing foreground images $O_{fg}$ while preserving the desired style characteristics.

\subsection{Fusion Module}

The Fusion Module (FM) uses an encoder-decoder setup for generating text images with realistic backgrounds. The encoder has three downsampling blocks and one residual block, while the decoder includes three upsampling blocks (see Figure ~\ref{fig:whole_pipeline}).

To make the results more realistic, FM's decoder layers employ a concatenation technique. Each decoder layer combines its output with the output from the Text Removal Module (TRM) within the Background Inpainting Module (BIM) decoder layer. This way, the decoder layers can use the contextual information from TRM, improving the overall coherence and quality of the generated text images.

The loss function used in FM combines GAN loss and L2 loss to optimize the final outcome. GAN loss makes FM create text images indistinguishable from real images, as it pits the generator against a discriminator. The discriminator distinguishes real from generated images, while the FM aims to fool the discriminator, leading to more realistic and high-quality results. L2 loss (mean squared error) measures the pixel-wise difference between the generated text image and the actual ground-truth image. The L2 loss ensures the generated text images closely resemble the target, highlighting fine details and accurate pixel values. 

The L2 loss is

\begin{equation}
\begin{split}
    L_{f\_l2} = \|T_f-O_f\|_2
\end{split}
\end{equation}
Where $T_f$ represents the groundtruth of the final result, and $O_f$ is the result generated by our model.

The GAN loss is
\begin{equation}
\begin{split}
    L_{f\_gan} = \mathbb{E}\{\log D_f(T_f, O_{b}) + \log(1-D_f(O_f, O_{b})\}
\end{split}
\end{equation}

The loss function utilized for this module is defined as follows:

\begin{align}
    L_{f} = \lambda_{f1} L_{f\_l2} + L_{f\_gan}
\end{align}

Where $\lambda_{f1}$ is balance factor and is set to 10.

To make sure that the final image $O_f$ is readable, we adopted a recognizer loss that uses cross-entropy loss. This recognizer loss plays a crucial role in guiding the Fusion Module (FM) during training, encouraging the generated text images to be not only realistic but also recognizable and accurate in terms of the textual content. 

The recognizer loss is

\begin{align}
    L_{rec} =\frac{1}{N}\sum^{N}_{t=1}\log(p_i|g_i)
\end{align}

where $p_i$ and $g_i$ are the prediction and ground truth of the labels. N indicates the maximum length, which is the sum of lowercase and uppercase characters.
In pursuit of generating more realistic images, we have adopted the VGG-loss, a content loss introduced in the seminal paper "Perceptual Losses for Real-Time Style Transfer and Super-Resolution." This loss function finds application in super-resolution and style transfer tasks, providing a perceptually meaningful alternative to conventional pixel-wise losses. During training, the VGG-loss fosters the alignment of the generated images' high-level content representations with those of the reference images. This alignment proves instrumental in preserving the original content's structural and semantic information, a critical aspect for tasks such as style transfer and super-resolution. Notably, unlike pixel-wise losses, the VGG-loss incorporates perceptual similarity considerations, rendering it a more efficacious and purposeful loss function for these undertakings.
The VGG loss is

\begin{align}
    L_{vgg} = \lambda_{v1}L_{per} + \lambda_{v2}L_{style}
\end{align}

Where $\lambda_{v1}$ and $\lambda_{v2}$ are set to 1 and 500, respectively.

The whole loss function can be expressed as

\begin{align}
    L = L_{b\_i} + L_{ts} + L_{f} + \lambda_{1}L_{vgg} + \lambda_{2}L_{rec}
\end{align}

Where $\lambda_{1}$ and $\lambda_{2}$ are set to 1 and 0.1 respectively.

\section{Experiments}

\subsection{Setup}
\subsubsection{Implementation Details}
We follow~\cite{qu2023exploring}, creating synthetic pairwise images from a diverse dataset of 400+ fonts and 10,000 backgrounds via SRNet Datagen~\cite{githubGitHubYoudaoaiSRNetDatagen}. The training set includes 1 million images, resized to 64 × 256 pixels, with a batch size of 16. Optimization uses Adam~\cite{kingma2014adam} ($\beta_1$ = 0.9, $\beta_2$ = 0.999) in PyTorch~\cite{paszke2019pytorch}, training for 800,000 iterations for efficiency and high-quality results.

\subsubsection{Benchmark Dataset}
To evaluate the efficacy of our proposed method \model, we conducted extensive assessments using a synthetic dataset. This dataset was generated using the publicly available SRNet Datagen ~\cite{githubGitHubYoudaoaiSRNetDatagen}, enabling us to create a diverse and representative collection ideal for evaluating our method's performance. Moreover, we performed a benchmark comparison against the data generated by our own dataset generator. This comprehensive evaluation allows us to validate the effectiveness of our approach and assess its capabilities in relation to existing datasets.

\subsubsection{Evaluation Metrics}
We adopt the commonly used metrics in scene text editing to evaluate our proposed method.
\begin{itemize}
    \item MSE or Mean Square Error, also known as l2 error
    \item PSNR or Peak Signal Noise Ratio, which computes the ratio of peak signal to noise
    \item SSIM or Structural Similarity Index Measurement ~\cite{wang2004image}, which computes the mean structural similarity index between two images
    \item FID or Fréchet Inception Distance ~\cite{heusel2017gans}, which represent the distance of image in vector representations
    \item Word Recognition Accuracy, which evaluates word recognition accuracy by comparing the original target word with the word recognized by a text recognizer using our generated results, as outlined in \cite{baek2019wrong}.
\end{itemize}

\subsection{Results}
\subsubsection{Quantitative Results}
In Table ~\ref{tab:result}, we give some quantitative results of our method and the other two competing methods. Our proposed method \model~surpasses the other implementations in all metrics. The average l2 error decreased by over 0.005. The average PSNR increased by over 2.8, the average SSIM increased by over 0.12, the average FID decreased by over 2.7, and the average WRA increased by 6\% than the second-best method. Lower MSE, higher PSNR, higher SSIM, and lower FID represent how similar the generated result is compared to the ground truth.
\begin{table}[htb]
    \centering
    \tabcolsep=0.15cm
    \begin{tabular}{|l|ccccc|}
        \hline
         Method & MSE~$\downarrow$ & PSNR~$\uparrow$ & SSIM~$\uparrow$ & FID~$\downarrow$ & WRA~$\uparrow$\\
         \hline
         SRNet & 0.0133 & 20.069 & 0.687 & 25.870 & 26.9\\
         MOSTEL & 0.0103 & 21.709 & 0.727 & 16.407 & 39.58\\
         \model~& \textbf{0.0051} & \textbf{24.525} & \textbf{0.840} & \textbf{13.73} & \textbf{45.98}\\
         \hline
    \end{tabular}
    \caption{Comparison between State of The Art Scene Text Editing and Proposed method ~\model~on benchmark dataset SRNet Datagen}
    \label{tab:result}
\end{table}

Table ~\ref{tab:result_own_datagen} presents a comparative analysis between our proposed model and the two other competing methods on our own benchmark dataset, which was generated using our custom data generator. 
\begin{table}[htb]
    \centering
    \tabcolsep=0.15cm
    \begin{tabular}{|l|ccccc|}
        \hline
         Method & MSE~$\downarrow$ & PSNR~$\uparrow$ & SSIM~$\uparrow$ & FID~$\downarrow$ & WRA~$\uparrow$ \\
         \hline
         SRNet & 0.022 & 17.415 & 0.686 & 47.701 & 8.44 \\
         MOSTEL & 0.018 & 18.392 & 0.711 & 37.649 & \textbf{36.56}\\
         \model~& \textbf{0.016} & \textbf{18.965} & \textbf{0.730} & \textbf{29.683} & 33.14\\
         \hline
    \end{tabular}
    \caption{Comparison between State of The Art Scene Text Editing and Proposed method ~\model~on benchmark dataset OURS.}
    \label{tab:result_own_datagen}
\end{table}

Our model outperforms all competitors on the benchmark datasets, except slightly 3\% below in WRA metric on our own datagen benchmark, highlighting its effectiveness.

\subsubsection{Qualitative Results}
\begin{figure*}[h]
    \centering
  \includegraphics[width=\textwidth,height=6cm,width=16cm]{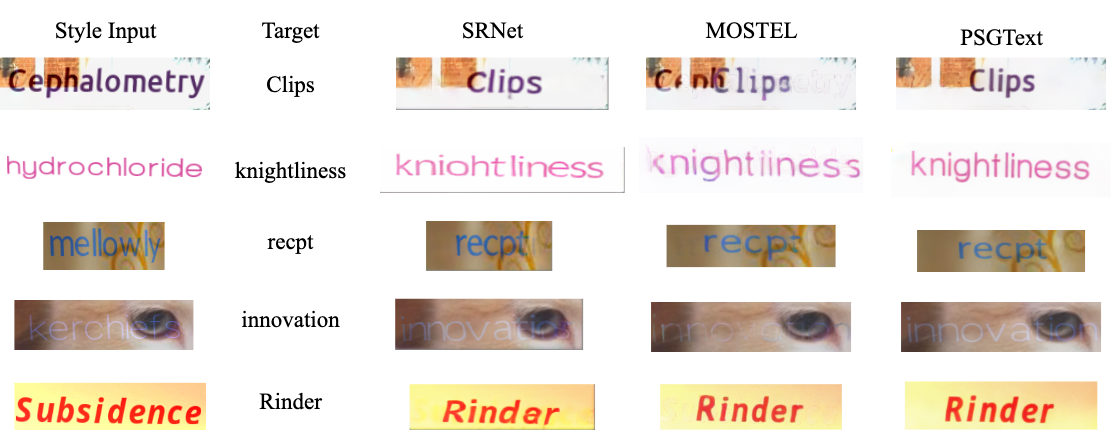}
  \caption{A qualitative comparison shows that our model excels in maintaining the original font style, including its unique characteristics and coloration. As evidenced in the third example of the third row, our model successfully transfers the style of cut text, a task that remains challenging for other existing methods. Furthermore, the fourth example in the same row showcases our method's proficiency in handling complex backgrounds, effectively preserving text while other approaches struggle to fully remove the original text.}
  \label{fig:qualitative_comparison}
\end{figure*}

In comparison to the previous works SRNet~\cite{wu2019editing} and MOSTEL~\cite{qu2023exploring}, our proposed method exhibits superior results, as demonstrated in Figure~\ref{fig:qualitative_comparison}. Notably, our model excels in maintaining the original font style, including its unique characteristics and coloration. As evidenced in the third example of the third row, our model successfully transfers the style of cut text, a task that remains challenging for other existing methods. Furthermore, the fourth example in the same row showcases our method's proficiency in handling complex backgrounds, effectively preserving text while other approaches struggle to fully remove the original text. These examples provide compelling evidence of the enhanced capabilities and robustness of our proposed method in various challenging scenarios.

\subsection{Ablation Studies}
In this section, we present the results of our ablation study conducted on our benchmark dataset to demonstrate the effectiveness of the modules within our proposed \model. Through systematic analysis, we examine the impact of individual components by selectively removing them and observing the performance changes. Additionally, we introduce another experiment that involves incorporating mask multiplication, inspired by the approach used in MOSTEL, to further validate the robustness and generalization of our model. The results from both the ablation study and the mask multiplication experiment provide valuable insights into the contributions of different components and their collective impact on the overall performance of our model.

\subsubsection{Pyramid Scene Parsing}
\begin{figure}[h]
  \includegraphics[width=8cm]{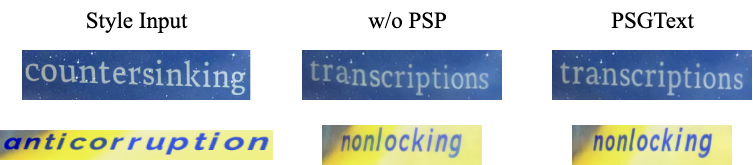}
  \caption{Comparison w and w/o PSP}
  \label{fig:ablation_PSP}
\end{figure}

The Pyramid Scene Parsing (PSP) module plays a pivotal role in capturing multi-scale contextual information, which is paramount for enhancing the performance of semantic image segmentation tasks. Upon removing the PSP module from our scene text editing architecture, we observed certain limitations that adversely impacted the quality of the generated images. Notably, in some cases, the model struggled to produce images with visually appealing characteristics. For instance, as depicted in Figure ~\ref{fig:ablation_PSP}, the scene text image generated with the PSP module exhibited a straight style, whereas the one without PSP exhibited a slightly bent appearance, which is deemed less desirable. Another salient issue observed in the absence of the PSP module was related to text color opacity. As illustrated in Figure ~\ref{fig:ablation_PSP}, the scene text image with the PSP module exhibited a more distinct color contrast in comparison to the one without PSP. Quantitatively, the results corroborated these observations, as shown in Table ~\ref{tab:ablation_studies_result}, where the model without the PSP module exhibited inferior performance compared to its PSP-equipped counterpart. These findings substantiate the indispensable role of the PSP module in our scene text editing architecture, affirming its efficacy in capturing essential contextual information and contributing to the generation of superior-quality images.

\subsubsection{Recognizer Loss}
In our ablation study, we investigated the impact of incorporating a recognizer loss during the training of our scene text editing model, aiming to improve image generation. This loss leverages a text recognition module that operates in tandem with the scene text editing process, encouraging recognizable and coherent text in edited scenes. This approach not only enhances the model's ability to accurately edit and manipulate text in the scene but also preserves legibility and natural appearance. Our results, as demonstrated in Table \ref{tab:ablation_studies_result}, show that this loss function improves overall image quality, including better fidelity in edited text, reduced artifacts, and improved preservation of contextual integrity. Prioritizing text recognizability instills a higher level of coherence and visual consistency in the generated images, aligning with the goal of producing easily interpretable and realistic edited text.

\subsubsection{Mask Multiplication}
Inspired by ~\cite{qu2023exploring}, we conducted an investigation into the integration of a mask multiplication technique into our model. This approach involves the multiplication of the inverted mask $O_{b\_mask}$ with the source image $I_s$, resulting in an image that exclusively contains the foreground text. The primary motivation behind this technique was to enable the text-swapping module to rapidly acquire style transfer capabilities.
However, as presented in Table ~\ref{tab:ablation_studies_result}, the results indicated that the addition of mask multiplication did not lead to improved outcomes. We hypothesize that this lack of improvement can be attributed to the limitations of the background inpainting module in generating accurate masks.

\begin{figure}
    \centering
    \includegraphics[width=\textwidth,width=6cm]{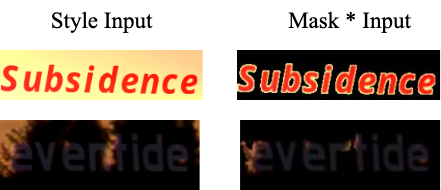}
    \caption{Mask Multiplication. In the first row, we observe a favorable outcome where the generated masks correctly isolate the foreground text. However, in the second row, the inaccurate masks lead to an incorrect combination of foreground and background text, resulting in a failed style transfer. }
    \label{fig:mask_multiplication}
\end{figure}

Figure ~\ref{fig:mask_multiplication} demonstrates the visual representation of the mask multiplication technique, illustrating both successful and unsuccessful cases. In the first row, we observe a favorable outcome where the generated masks correctly isolate the foreground text. However, in the second row, the inaccurate masks lead to an incorrect combination of foreground and background text, resulting in a failed style transfer. Consequently, the failure to produce appropriate foreground source images potentially confounded the text-swapping module during the style transfer process.

 \subsubsection{Skeleton Loss}
Drawing inspiration from the work of SRNet \cite{wu2019editing}, we embarked on an investigation into the integration of skeleton loss within our model. This strategic inclusion aimed to expedite the refinement process of the Text Swapping Module, leading to the generation of an improved output denoted as $O_{fg}$. The outcomes, as outlined in Table \ref{tab:ablation_studies_result}, distinctly showcase the efficacy of incorporating the skeleton loss, resulting in notable performance enhancements.

\begin{table}[htb]
    \centering
    \tabcolsep=0.15cm
    \begin{tabular}{|l|cccc|}
        \hline
         Method & MSE~$\downarrow$ & PSNR~$\uparrow$ & SSIM~$\uparrow$ & FID~$\downarrow$ \\
         \hline
          SRNet & 0.0133 & 20.069 & 0.687 & 25.870 \\
         MOSTEL & 0.0103 & 21.709 & 0.727 & 16.407 \\
         \model~w/o PSP & 0.0064 & 23.56 & 0.801 & 15.84 \\
         \model~w/o rec & 0.0059 & 23.899 & 0.822 & 14.864 \\
         \model~w/ MM & 0.0061 & 23.395 & 0.834 & 17.603 \\
         \model~w/o sk & 0.0068 & 23.148 & 0.794 & 17.827 \\
         \model~& \textbf{0.0051} & \textbf{24.525} & \textbf{0.840} & \textbf{13.73} \\
         \hline
    \end{tabular}
    \caption{The ablation study of \model~model architecture. Three essential components: Pyramid Scene Parsing, Recognizer Loss and Mask Multiplication are removed/added to demonstrate its effectiveness.}
    \label{tab:ablation_studies_result}
\end{table}

\subsection{Limitation}
\begin{figure}[thb]
    \centering
    \includegraphics[width=6cm]{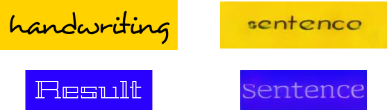}
    \caption{Failure cases of \model~when dealing with images featuring intricate structures or utilizing rare font shapes. The image on the left represents the input, while the image on the right shows the model's output with the text transformed to "sentence" text.}
    \label{fig:failure_case}
\end{figure}

While our method demonstrates capability in handling most scene text images, it is important to acknowledge certain limitations. Specifically, \model~may encounter difficulties when attempting to modify text in scene text images that exhibit highly intricate structures or employ rare font shapes. These limitations signify areas for further improvement and optimization in our methodology. Figure~\ref{fig:failure_case} shows some failed cases of \model. 

\section{Conclusion}
In this study, we introduced \model, a sophisticated end-to-end trainable deep neural network tailored for scene text editing. This model adeptly replaces text within images through a nuanced process of style and content decomposition, ensuring the preservation of original elements during style-content transfers. Importantly, \model~has exhibited superior performance compared to prior methods on established benchmarks. Our rigorous testing unambiguously highlights the model's exceptional precision in scene text editing, maintaining high accuracy and visual integrity, thus affirming its effectiveness in addressing a wide range of text editing tasks.

\clearpage
\appendix


\bibliography{aaai24}

\end{document}